# Back to Bytes: Revisiting Tokenization Through `UTF-8`

**Amit Moryossef**[β]    **Clara Meister**[B]    **Pavel Stepachev**[Θ]    **Desmond Elliott**[o]
[β]`sign.mt`    [B]ETH Zürich    [Θ]University of Edinburgh    [o]University of Copenhagen
amit@sign.mt  meistecl@inf.ethz.ch  de@di.ku.dk

## Abstract

We present `UTF8Tokenizer`, a minimalist byte-level tokenizer that maps text *exactly* to IDs corresponding to the bytes underlying the text's `UTF-8` encoding (e.g., byte \x09 is token ID 9). Unlike prior byte-level approaches (Xue et al., 2021; Pagnoni et al., 2025), our implementation *never* introduces out-of-range IDs (i.e. there is no token ID 256) or auxiliary tokens: all special behavior (e.g. padding, boundaries, conversation structure, attention segments, tool calling, "thinking" spans, etc.) is encoded using `C0` control bytes—just as ASCII was originally designed to embed control information alongside printable text. These design principles yield practical benefits: (1) faster tokenization (14×) and significantly lower host–device transfer (8× less than `int64`); (2) simple, shareable $256 \times d$ embedding tables that can be aligned across models; and (3) a training-time enhancement via *bit-biased embeddings*, which exposes per-byte bit structure and can be added to the embedding table post-training, removing inference costs. Our HuggingFace-compatible implementation improves language modeling convergence.[1]

## 1 Introduction

Tokenization specifies how raw text becomes model-consumable inputs. Multiple tokenization paradigms exist: most prominently subword and byte-level schemes. Byte-level tokenization maps text, under a fixed character encoding (e.g., `UTF-8`), to tokens that correspond directly to bytes. Historically, models utilizing byte-level tokenizers have performed worse than their subword-level counterparts in terms of performance and efficiency, but recent advances have closed much of this gap, making byte-level modeling a compelling alternative (Pagnoni et al., 2025). Notably, byte-level tokenization still requires mapping text to representations processable by a language model, typically achieved by converting text to token IDs and then to model-consumable vectors via a learned embedding lookup table. Most byte-level tokenizers (e.g. Xue et al., 2021; Pagnoni et al., 2025) implement this step using special conventions developed in the context of subword-level tokenization schemes, such as expanding the vocabulary with additional IDs for special tokens. These conventions reintroduce complexity and overlook the structural regularities of encoding schemes (e.g., leading-byte patterns and continuation bytes in `UTF-8`) that can be exploited without introducing extra symbols.

In this work, we revisit byte-level tokenization conventions. We propose a simple scheme that follows an uncompromising constraint: *never leave the 0–255 token ID range*. The `UTF8Tokenizer` defines the token sequence as exactly the input's `UTF-8` byte sequence encoding, with special conventions represented by repurposed `C0` control bytes. This enables a fixed, universal vocabulary that can be defined exactly within a single $256 \times d$ embedding table. Memory-wise, we store tokens as `uint8`, enabling zero-copy views[2] and minimizing data movement (§3.3). We further propose a lightweight bit-bias that can be added to token embeddings. This bias exposes structural relationships between bytes—such as the aforementioned leading byte patterns—which enables more efficient learning for language models. Our new bytes-only tokenizer highlights the simplicity of byte-based inputs while exploiting structural information intrinsic to Unicode.

**Contributions.**

1. **Strict bytes-only tokenizer.** A drop-in HuggingFace-compatible tokenizer that *only*

---

[1]Our tokenizer, code, and experiments are available at https://github.com/sign/utf8-tokenizer.

[2]Token sequences can be viewed directly as contiguous memory, avoiding expensive data copies or conversions.

uses 0–255 IDs, with exact `UTF-8` tokenization/detokenization and control-byte semantics for structure.

2. **Control-token protocol.** A practical, expandable protocol for special tokens such as padding, string/message boundaries, attention segments, tool calls, and "thinking" spans.

3. **Systems gains.** Empirical efficiency from `uint8` tokens: faster tokenization, lower memory traffic, and reduced host–device transfer.

4. **Bit-biased embeddings.** A lightweight training-time augmentation that exposes per-byte bit patterns; with zero cost at inference.

## 2 Tokenization

We follow standard Unicode terminology. Text is a sequence of code points in the Unicode codespace; a *Unicode Transformation Format* (UTF) such as `UTF-8` serializes those code points into a sequence of bytes. A tokenizer maps text to token IDs; these IDs index a learned embedding table, from which we gather the vectors that are ultimately given to a language model as input.

Tokenization can be done at both the subword and byte level—the difference is what constitutes a token. The choice induces different modeling biases, shaping efficiency and cross-lingual behavior.

**Subword-level Tokenization.** Most modern models rely on popular subword methods such as Byte-Pair Encoding (BPE; Sennrich et al. (2016)) and UnigramLM (Kudo, 2018). These methods introduce powerful inductive biases that, for many widely used languages, are helpful for language modeling. Yet, for a large number of the world's other languages, these methods lead to disparate segmentation lengths and non-sensical token boundaries (Foroutan et al., 2025), which is a source of disparities in model costs and performance (Petrov et al., 2023; Ahia et al., 2023; Lesci et al., 2025). These disparities stem partly from differences in resource availability across languages but also from the algorithmic heuristics in subword algorithms that are not universally applicable.

**Byte-level Tokenization.** Byte-level tokenization (Xue et al., 2021; Pagnoni et al., 2025) fixes token boundaries at the level of encoded bytes: given a chosen character encoding (e.g., `UTF-8`), the tokens are exactly the resulting byte values. On the downside, byte-level tokenization abandons helpful inductive biases given by subword vocabularies, where frequent morphemes/words are short token sequences. In this case, a model needs to learn lexical and morphological regularities itself. Arguably, the advantages of byte-level approaches outweigh this: by letting the model learn textual units implicitly, in a more end-to-end manner, they remove learned vocabularies and language-specific segmentation rules, ensuring uniform vocabulary size across languages and robustness to mixed-script and rare characters.

## 3 `UTF8Tokenizer`

We model tokenization as identity on bytes. No normalization, pretokenization,[3] or learned vocabulary is required. Every valid `UTF-8` bytes string is exactly representable and reversible.

```
def tokenize(text: str) -> list[int]:
    return list(text.encode("utf-8"))

def detokenize(tokens: list[int]) -> str:
    return bytes(tokens).decode("utf-8")
```

**IDs and storage.** We implement a HuggingFace-compatible tokenizer, where token IDs are the byte values themselves (0–255), stored as `uint8`. This cuts memory requirements by $8\times$ compared to the standard `int64`, reduces host–device transfer, and allows for zero-copy views. Embedding tables are a single $256 \times d$ matrix.

### 3.1 Control-Token Protocol

To encode structural information while preserving the strict 0–255 byte range, we repurpose the `C0` control characters as special tokens (Table 1). These tokens are valid bytes, but should not occur in natural text, making them a convenient substrate for task-specific semantics. Our protocol covers padding, text boundaries, conversation turns, attention regions, private reasoning, and tool usage.

**Padding:** `<NUL>` (Null; 00) is the padding symbol. It never appears in unpadded spans, allowing models to safely mask on this token.

**String boundaries:** Every sequence can be wrapped with `<STX>` (Start of Text; 02) and `<ETX>` (End of Text; 03), providing BOS/EOS semantics.

---

[3]*Normalization* refers to rewrites of Unicode text prior to tokenization (e.g., decomposing diacritics or folding ligatures). *Pretokenization* refers to any text-dependent segmentation performed before token–ID mapping (e.g., whitespace/punctuation splitting or language-specific segmentation).

| Dec | Hex | Abbr. | Name | Description | `UTF8Tokenizer` Usage |
|---|---|---|---|---|---|
| 0 | 00 | NUL | Null | Does nothing; padding/blank tape | Padding |
| 1 | 01 | SOH | Start of Heading | Begin heading of message | Begin message block |
| 2 | 02 | STX | Start of Text | Start of message text | Beginning of string |
| 3 | 03 | ETX | End of Text | End of message text | End of string |
| 4 | 04 | EOT | End of Transmission | End of transmission | – |
| 5 | 05 | ENQ | Enquiry | Trigger a response | Begin thinking |
| 6 | 06 | ACK | Acknowledge | Indicate successful receipt | End thinking |
| 7 | 07 | BEL | Bell, Alert | Call for attention | – |
| 8 | 08 | BS | Backspace | Move left | – |
| 9 | 09 | HT | Horizontal Tab | Move to next tab stop | Whitespace |
| 10 | 0A | LF | Line Feed | Move down to next line | Whitespace |
| 11 | 0B | VT | Vertical Tab | Move down to next vertical tab stop | Whitespace |
| 12 | 0C | FF | Form Feed | Move to top of next page | Whitespace |
| 13 | 0D | CR | Carriage Return | Move to column zero | Whitespace |
| 14 | 0E | SO | Shift Out | Switch to alternative character set | Begin attention block |
| 15 | 0F | SI | Shift In | Return to regular character set | End attention block |
| 16 | 10 | DLE | Data Link Escape | Interpret following chars differently | – |
| 17 | 11 | DC1/XON | Device Control 1 | Flow control | Tool Definition |
| 18 | 12 | DC2 | Device Control 2 | Device control | – |
| 19 | 13 | DC3/XOFF | Device Control 3 | Flow control | – |
| 20 | 14 | DC4 | Device Control 4 | Device control | – |
| 21 | 15 | NAK | Negative Acknowledge | Negative response | – |
| 22 | 16 | SYN | Synchronous Idle | Idle state | – |
| 23 | 17 | ETB | End of Transmission Block | Marks end of data block | End of message block |
| 24 | 18 | CAN | Cancel | Invalidate preceding data | – |
| 25 | 19 | EM | End of Medium | End of usable medium | – |
| 26 | 1A | SUB | Substitute | Replace invalid character | Begin tool calling |
| 27 | 1B | ESC | Escape | Alter meaning of following bytes | End tool calling |
| 28 | 1C | FS | File Separator | Field delimiter | – |
| 29 | 1D | GS | Group Separator | Group delimiter | – |
| 30 | 1E | RS | Record Separator | Record delimiter | – |
| 31 | 1F | US | Unit Separator | Unit delimiter | – |
| 127 | 7F | DEL | Delete | Delete character | – |

Table 1: Full list of `C0` control characters and their repurposed roles in the `UTF8Tokenizer`. Bytes `09-0D` are whitespace characters, normally used in text, and should not be repurposed.

**Messages Blocks:** Instruction-tuning messages can be surrounded by <SOH> (Start of Heading; 01) and <ETB> (End of Transmission Block; 04).

**Attention regions:** Segments intended for full self-attention—such as PrefixLM (Raffel et al., 2019) or MAS (Katz et al., 2025)—can be enclosed in <SO> (Shift Out; 14) and <SI> (Shift In; 15).

**Thinking spans:** Private "chain-of-thought" reasoning can be bracketed by <ENQ> (Enquiry; 05) and <ACK> (Acknowledge; 06).

**Tool calls:** External calls can be wrapped with <SUB> (Substitute; 26) and <ESC> (Escape; 27).

**Extensibility:** The remaining `C0` positions are intentionally left unassigned, allowing researchers and practitioners to introduce additional control semantics without altering the tokenizer vocabulary.

### 3.2 Visualization and Safety

Control bytes are invisible in plain text. We recommend using Unicode Control Pictures (U+2400–U+2426) when *displaying* bytes for debugging/data reviews, so logs/UI are legible while the underlying byte stream remains unchanged. Appendix A provides LaTeX macros, and our library provides a `visualize_control_tokens` utility.

### 3.3 Zero-Copy and Transfer

Representing tokens as `uint8` reduces the data volume of a sequence by $8\times$ relative to `int64`.

In many frameworks and programming languages, such as NumPy (Harris et al., 2020) or PyTorch (Paszke et al., 2019) we can create zero-copy views, reducing both memory and compute:

```
ids = text.encode("utf-8")

import numpy as np
np.frombuffer(ids, dtype=np.uint8)

import torch
torch.frombuffer(ids, dtype=torch.uint8)
```

In our benchmarks, this proves to be $14\times$ faster compared to the `ByT5Tokenizer` in Transformers.

## 3.4 Example Text Representation

Below is an example input to a `UTF8Tokenizer`-based language model that features multiple message blocks (`<SOH>`/`<ETB>`), MAS attention patterns (`<SO>`/`<SI>`), a thinking span (`<ENQ>`/`<ACK>`), and a tool call (`<SUB>`/`<ESC>`).

> `<STX><SOH>`system
>
> `<SO>`You are a helpful assistant`<SI><ETB>`
>
> `<SOH>`user
>
> `<SO>`How much is 1+2?`<SI><ETB>`
>
> `<SOH>`assistant
>
> First I'll think about it.
>
> `<ENQ>`The user wants me to calculate, I should call the calculator `<SUB>`{"type": "calculator", "expression": "1+2"}`<ESC>`3`<ACK>`
>
> 1 + 2 = 3`<ETB><ETX><NUL><NUL><NUL><NUL>`

For instruction tuning with your data, use `apply_chat_template` in our tokenizer.

## 4 Bit-Biased Byte Embeddings

Byte IDs hide simple regularities inherent in Unicode's binary structure: digits 0-9 share the same high nibble (upper four bits), and in scripts like Latin, Cyrillic, Greek, and Armenian, letter case differs only by bit 5. These structured relationships are lost when bytes are treated as opaque categorical IDs. We hypothesise that making binary regularities explicit improves sample efficiency and stabilizes optimization, especially for small models trained on bytes. To this end, we propose exposing this structure by adding a small linear projection of the 8 bit features of each token to its embedding. Concretely, for token $t \in [0, 255]$ we compute[4] $h(t) \in \{0,1\}^8$ and learn a small embedding matrix $W_{bit} \in \mathbb{R}^{8 \times d}$. We then use $\text{embed}(t) = E[t] + h(t)W_{bit}$ as the embedding for token $t$. This only adds $8d$ parameters.

```python
def unpack_bits(x: ByteTensor) -> ByteTensor:
    shifts = torch.arange(7, -1, -1)
    return (x.unsqueeze(-1) >> shifts) & 1

def embed(tokens: ByteTensor) -> FloatTensor:
    bits = unpack_bits(tokens)  # (B, L, 8)
    bit_feats = bits @ W_bit    # (B, L, D)
    emb = E[tokens]             # (B, L, D)
    return emb + bit_feats
```

At inference, we can fold $W_{bit}$ into $E$, by setting $E' = E + [h(0), \ldots, h(255)] \cdot W_{bit}$ as our embedding. Thus, we still work with a standard $256 \times d$ table, leaving latency and memory unchanged.

---

[4]For $h(t)$, we use a most significant bit to least significant bit vector representation, i.e., $h(t)_k = \lfloor \frac{t}{2^{7-k}} \rfloor \mod 2$.

## 5 Experiments

**Setup.** We adapt the language modeling example[5] on HuggingFace Transformers (Wolf et al., 2019) and train a small decoder-only LM (Llama; Touvron et al., 2023) with 4 layers, 4 attention-heads, intermediate-size of 640, and hidden-size of 256 (following `sbintuitions/tiny-lm`) on Wiki-Text (`wikitext-2-raw-v1`) with byte-level tokenization. We compare vanilla byte embeddings vs. bit-biased (+$8d$ params during training). For an exact comparison with ByT5 (Xue et al., 2021), we add a BOS token to their tokenizer, to result in the same sequence length.

**Results.** Table 2 shows the results of training with the `UTF8Tokenizer` and the bit-biased embeddings. Bit-biased embeddings decrease validation loss and increase byte-accuracy strictly on all evaluation steps. At inference, folding $W_{bit}$ into $E$ maintains identical $256 \times d$ parameters and latency.

| Tokenizer | Perplexity ↓ | Accuracy ↑ |
|---|---|---|
| ByT5 (2021) | $1.957 \pm 0.004$ | $0.451 \pm 0.002$ |
| UTF8Tokenizer | $1.947 \pm 0.018$ | $0.451 \pm 0.003$ |
| + bit-bias | **$1.940 \pm 0.014$** | **$0.454 \pm 0.003$** |

Table 2: Language model metrics for different byte-level tokenizers, on the evaluation set, trained for 1 epoch on `wikitext-2-raw-v1`, with 3 seeds.

## 6 Practical Guidance

**Drop-in adoption.** Replace your tokenizer with `UTF8Tokenizer` and set all special IDs to the corresponding C0 bytes (e.g., `pad_token_id=0`, `bos_token_id=2`, `eos_token_id=3`). No vocabulary files or merges are needed.

**Data hygiene.** If your corpus contains C0 bytes with semantic meaning, our protocol will interpret them as structure. In such rare cases, escape the content at the *application* layer; avoid inventing new token IDs to preserve the bytes-only contract.

**Instruction tuning.** Wrap each message with `<SOH>`/`<ETB>`, mark private reasoning with `<ENQ>`/`<ACK>` (optionally filtered at decode), and bracket tool calls with `<SUB>`/`<ESC>`. Prefix-LM regions can be annotated with `<SO>`/`<SI>`.

---

[5]Full training details are available in our repository under `experiments/language-modelling`

**Bit-Bias.** We suggest applying bit-biased embeddings as a temporary training-time augmentation rather than a permanent architectural change. Concretely, implement a training callback that patches the embedding layer with the bit-projection map. Monitor the gradient norm of this projection; once it becomes small, remove the patch and continue training with the folded $256 \times d$ embedding table. This procedure yields the best of both worlds: during early training, bytes that share structural regularities, are encouraged to align, while later training proceeds with standard embeddings and no additional computational overhead. In effect, this acts as a data-aware embedding initialization strategy that can improve convergence without altering the final model architecture.

## 7 Control Safety and ASCII Isolation

All `ASCII` bytes (`0x00–0x7F`), including our repurposed control tokens, can only appear as single-byte code points in `UTF-8`. They are never part of any multi-byte sequence. This strict separation ensures full backward compatibility with plain `ASCII`. Consequently, our control tokens are unambiguous (i.e., we can always stop decoding when we see 0x03), and the Latin alphabet enjoys a dedicated, non-overlapping allocation in the embedding matrix: 26 uppercase and 26 lowercase letters that cannot collide with any other Unicode script. Using bit-biased embeddings, information is shared across scripts through the $W_{bit}$ projection.

## 8 Conclusion

We introduce `UTF8Tokenizer`, a strict bytes-only tokenizer with a practical control-byte protocol and a training-time bit-biasing enhancement that folds away at inference. The approach simplifies multilingual NLP pipelines, reduces memory traffic, and accelerates tokenization. We advocate strict byte modeling as a robust default for efficiency-centric and multilingual settings.

## 9 Limitations

In rare corpora that contain meaningful C0 bytes, our control protocol (particularly `<NUL>`, `<STX>`, and `<ETX>`) may collide unless content is isolated or escaped upstream.

Land and Bartolo (2024) note in *Appendix C* that certain byte values almost never appear in `UTF-8` text and are therefore untrained in language models. These include the invalid leading bytes `0xF5–0xFF` and `0xC0–0xC3`, and to a lesser extent the rarely used `0xF1`, `0xF2`, and `0xF4`. Beware if—like Hwang et al. (2025)—you decide to use these as additional special tokens, since they will need to be stripped away before decoding the bytes to text.

## Acknowledgements

We thank Tiago Pimentel, Alex Warstadt, and Ethan Wilcox for early discussions, and particularly for proposing the elegant folding of the bit-biased embeddings back to the original matrix.

## A LaTeX Macros

When writing about control tokens in LaTeX, typesetting them using control tokens is non-trivial. To simplify this, we recommend defining the following macros for convenient shortcuts for frequently used tokens:

```latex
\usepackage[dvipsnames]{xcolor}

% Define token macros (uppercase names)
\newcommand{\token}[2]
    {\textcolor{#1}{\texttt{<#2>}}}
\newcommand{\tokenNUL}{\token{Gray}{NUL}}
\newcommand{\tokenSTX}{\token{ForestGreen}{STX}}
\newcommand{\tokenETX}{\token{ForestGreen}{ETX}}
\newcommand{\tokenSOH}{\token{Violet}{SOH}}
\newcommand{\tokenETB}{\token{Violet}{ETB}}
\newcommand{\tokenSO}{\token{Red}{SO}}
\newcommand{\tokenSI}{\token{Red}{SI}}
\newcommand{\tokenENQ}{\token{Orange}{ENQ}}
\newcommand{\tokenACK}{\token{Orange}{ACK}}
\newcommand{\tokenSUB}{\token{Brown}{SUB}}
\newcommand{\tokenESC}{\token{Brown}{ESC}}
```